\documentclass[runningheads]{llncs}
\usepackage{graphicx}
\usepackage{tikz}
\usepackage{comment}
\usepackage{amsmath,amssymb} % define this before the line numbering.
\usepackage{color}

\usepackage[accsupp]{axessibility}  % Improves PDF readability for those with disabilities.

\usepackage{multirow}
\usepackage{booktabs}
\usepackage{xcolor}

\begin{document}
% \linenumbers
\pagestyle{headings}
\mainmatter
\def\ECCVSubNumber{}  % Insert your submission number here

\title{Distinctive Image Captioning via CLIP Guided Group Optimization} % Replace with your title

\begin{comment}
\titlerunning{ECCV-22 submission ID \ECCVSubNumber} 
\authorrunning{ECCV-22 submission ID \ECCVSubNumber} 
\author{}
\institute{}

\end{comment}
%******************

% \begin{comment}
\titlerunning{Abbreviated paper title}
\author{Youyuan~Zhang \inst{2} \and
        Jiuniu~Wang \inst{3} \and
        Hao~Wu \inst{4} \and
        Wenjia~Xu   \inst{1}\thanks{Corresponding author.} } 
\authorrunning{Y. Zhang et al.}
\institute{State Key Laboratory of Networking and Switching Technology, Beijing University of Posts and Telecommunications  \and McGill University \and
          Department of Computer Science, City University of Hong Kong\\
		  % \email{jiuniwang2-c@my.cityu.edu.hk} 
		   \and
		   South China University of Technology\\
		   \email{youyuan.zhang@mail.mcgill.ca} \quad
          \email{xuwenjia@bupt.edu.cn}
		  % \email{cth\_wu@mail.scut.edu.cn} 
		    }
% \end{comment}
%******************
\maketitle

\begin{abstract}

Image captioning models are usually trained according to human annotated ground-truth captions, which could generate accurate but generic captions. In this paper, we focus on generating distinctive captions that can distinguish the target image from other similar images. To evaluate the distinctiveness of captions, we introduce a series of metrics that use large-scale vision-language pre-training model CLIP to quantify the distinctiveness. To further improve the distinctiveness of captioning models, we propose a simple and effective training strategy that trains the model by comparing target image with similar image group and optimizing the group embedding gap. Extensive experiments are conducted on various baseline models to demonstrate the wide applicability of our strategy and the consistency of metric results with human evaluation. By comparing the performance of our best model with existing state-of-the-art models, we claim that our model achieves new state-of-the-art towards distinctiveness objective.

\keywords{Distinctive Image Captioning, CLIP, Similar Image Group, Group Embedding Gap}
\end{abstract}

\section{Introduction}

% \wenjia{What is image captioning, and why is it important.}

The task of image captioning involves interpreting the semantic information of images into natural language sentences. It is considered as a fundamental task in both computer vision and natural language processing. Image captioning is applied in various scenario such as assisting visually impaired people \cite{makav2019new}, search engine \cite{frankel1996webseer}, medical imaging \cite{xiong2019reinforced}, etc.

% \wenjia{What is distinctive image captioning? What did other works do towards distinctive image captioning? Why is this problem still important?}

General requirements of image captioning include two aspects: fluency, as the sentence should be well-formed as natural language; descriptiveness, as the sentence should accurately describe salient information of the image. In this paper, we focus on another important aspect: distinctiveness, which aims to generate captions with sufficient details and distinguish the target image from other similar ones \cite{wang2020compare}. Currently, most of the existing models only use human annotated ground-truth captions as labels. These models naturally focus on the similarity of generated captions and ground-truth captions. As a result, the generated captions are usually too generic because they are trained to be close to all ground-truth captions. Some pioneers \cite{shetty2017speaking,luo2018discriminability} introduce diversity and discriminability objective that operate on semantic embeddings of images to generate more diverse or discriminative captions. However, as discussed in \cite{wang2020compare}, models following these objectives generate grammarly different sentences with similar meanings, which does not follow the requirement of distinctiveness.

% \wenjia{Our general idea of solving this problem. Then explain in detail what we do.}
% \wenjia{I don't quite agree with this sentence. Human annotations usually include rich details.}
% \wenjia{Jiuniu's ACMMM paper already used similar image group, right?}. \wenjia{I would suggest change the restrictions into following three parts: 1. Only traditional language quality metrics are used for training. 2. the generated captions may not correctly describe the image content. 3. Some captions describing the semantic of images are not similar to the ground truth captions, thus get low cider score.}

Some of the restrictions of traditional training strategies that prevent models from being distinctive include: 1) Only traditional language quality metrics, such as CIDEr \cite{vedantam2015cider}, are used for training; 2) Ground-truth captions are annotated by human which does not necessarily include much details other than salient objects. Therefore, optimizing towards ground-truth captions through language quality metrics naturally makes captioning models focus on salient information and ignore distinctive details in images. To overcome the intrinsic shortcomings of traditional training strategy, we hereby propose a group based optimization strategy that uses CLIP \cite{radford2021learning} to guide distinctive training. We propose a series of reference-free metric based on CLIP to evaluate the relevance and distinctiveness of captions and images. Furthermore, we introduce similar group which is formed by the similar images of an target image. The similar group can be seen as hard negatives from which the model explicitly learns to distinguish detailed features. During training, our group based optimization takes the similarity of generated captions with target images as positive reward and takes the similarity with similar images as negative reward. The similarity is obtained from CLIP thus is not affected by the drawbacks of human annotations. Our proposed optimization strategy is ``plug and play" which can be applied on any sequence prediction model. Experimental results show that our proposed strategy achieves outstanding performance on a variety of existing models.

% \wenjia{Summarize the contribution of this work.}
% \wenjia{1. propose network 2. propose metric 3.  Extensive experiment results demonstrate the effectiveness of our model.}
To summarize, the contributions in this paper are three fold: 1) We propose a series of reference-free metrics to quantify the distinctiveness of image captioning. 2) We propose a group based optimization strategy that exploits the new metrics as reward to guide distinctive training. 3) We conduct extensive experiments on various state-of-the-art captioning models and demonstrate the effectiveness of our strategy.

\section{Related Work}

\subsection{Image Captioning}
% \wenjia{definition. Relate works. LSTM -RNN -transformer XE loss -> reinforcement loss...}
Deep learning based captioning models generally employ encoder-decoder framework. Early attempts encode images using CNN \cite{simonyan2014very,he2016deep} based feature extractor, and output captions through RNN \cite{cho2014learning} and LSTM \cite{hochreiter1997long} based decoder  \cite{vinyals2015show,karpathy2015deep,ma2015multimodal,mao2014deep}. Some works use object-level features extracted by Faster-RCNN \cite{ren2015faster} and improves the performance. Recent works introduce attention mechanism to encourage cross-modal interaction between the two modalities and mitigate the long-term dependency problem in LSTM \cite{cho2015describing}. For example, \cite{anderson2018bottom} applies bottom-up attention at object-level features and map salient visual features to output words via top-down attention. \cite{pan2020x} applies X-Linear attention to leverage higher-order intra and inter-modal interactions.

\subsection{Objectives for Image Captioning}

Sequence prediction models are typically trained with maximum likelihood estimation (MLE) objective by cross-entropy loss. However, as discussed in \cite{ranzato2015sequence}, this method causes the problem of exposure bias, which accumulates the error of caption generation at test time. To address this problem, \cite{bengio2015scheduled} introduces scheduled sampling to reduce the bias. \cite{ranzato2015sequence,rennie2017self} treat sequence generation as reinforcement learning (RL) problem. In specific, \cite{rennie2017self} proposes self-critical sequence training that optimizes the policy gradient of the model parameters using REINFORCE with baseline. However, as pointed out in \cite{dai2017contrastive,dai2017towards,wang2019describing} such approaches usually result in over generic captions. Therefore, different objectives, such as diversity and discriminability, are proposed to enrich the expressive capacity of captions. In specific, Div-Cap \cite{wang2019describing} propose to measure diversity by analyzing latent semantic space of image encoding. G-GAN \cite{dai2017towards} use conditional generative adversarial network (CGAN) \cite{mirza2014conditional} to improve diversity of captions. Disc-Cap \cite{luo2018discriminability} trains a retrieval model to guide discriminative captioning. In terms of distinctiveness objective, VisPara-Cap \cite{liu2019generating} employs a two-stage LSTM model and generates two captions, where the second caption is pharaphrased from the first caption and is more distinctive. CiderBtw-Cap \cite{wang2020compare} re-weights the ground-truth captions by comparing their distinctiveness and trains the model using a weighted loss. Gdis-Cap \cite{wang2021group} designs a group attention module that re-weights the attention features of images to highlight the distinctive features. CLIP-Cap \cite{cho2022fine} directly use CLIP \cite{radford2021learning} to compute relevance of image-caption pairs and generates more distinctive captions. Compared with these methods, our proposed method integrates the advantages of the following: 1) We do not rely on a specific model and therefore our method is more widely applicable. 2) We use reference-free reward to avoid the drawbacks of human annotation. 3) We use similar image group to better guide distinctive training via comparing.

% \wenjia{add more works, add details. We need one section about distinctive image captioning. Introduce how related works improve the distinctiveness of captions, and specify our difference with these works, highlight our novelty and contribution.}

\subsection{Metrics for distinctive image captioning}

Traditional metrics, such as CIDEr \cite{vedantam2015cider}, SPICE \cite{anderson2016spice}, BLEU \cite{papineni2002bleu}, ROUGE \cite{lin2004rouge} and MENTEOR \cite{banerjee2005meteor}, are used to measure the similarity between generated captions and ground-truth captions. In terms of distinctiveness metrics, \cite{wang2020towards} designs SPICE-U and introduces a notion of uniqueness over concepts in captions. \cite{wang2020compare} proposes CIDErBtw metric which reflects the distinctiveness by measuring the similarity of the target captions and captions of similar images. \cite{wang2021group} proposes DisWordRate metric which reflects the percentage of distinctive words in captions.

% \wenjia{Specify our difference with these works, highlight our novelty and contribution.}

\section{Methodology}
% \wenjia{introduce each part of this section}
% \wenjia{draw a model architecture. }
% \jimmycmd{Add the pipeline of image captioning.}
% \jimmycmd{Need high resolution for text.}

\begin{figure}[pbt]
    \centering
    \includegraphics[width=\linewidth]{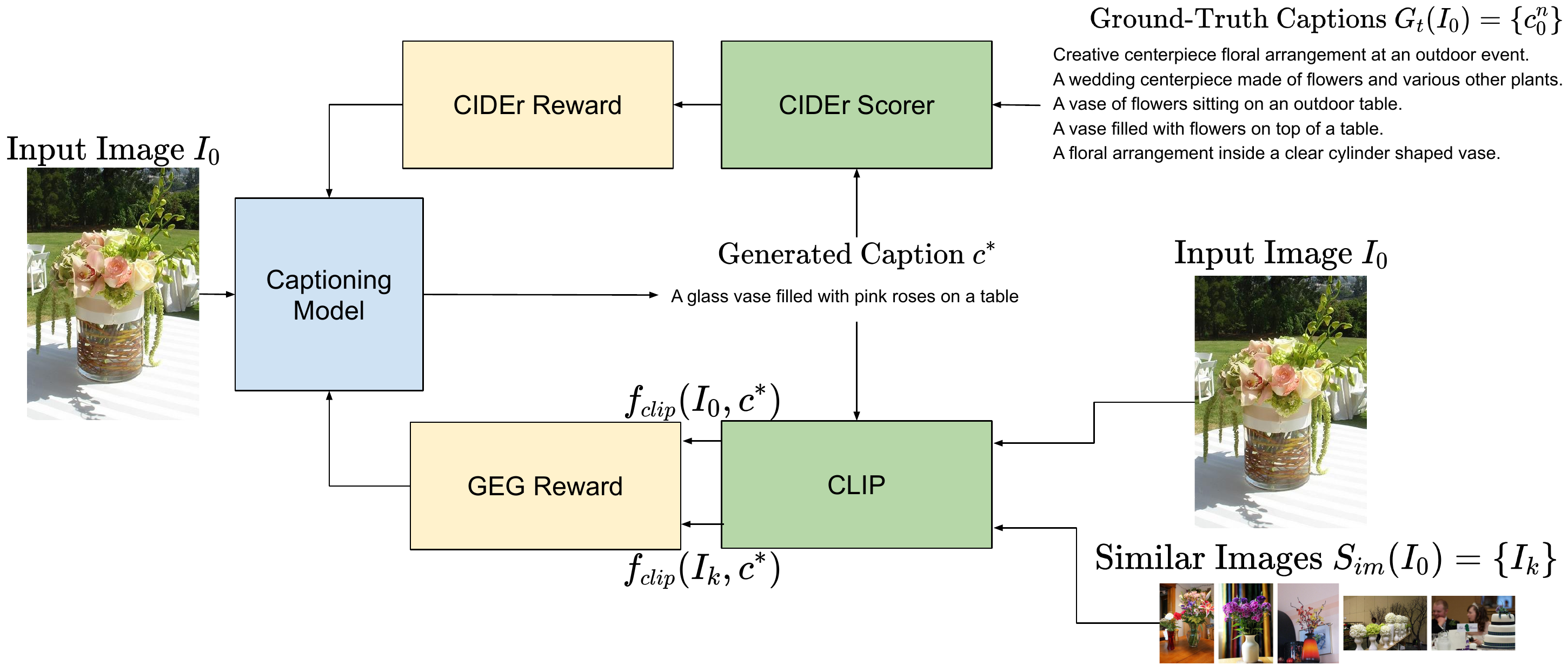}
    \caption{Pipeline of group optimization using weighted group embedding gap reward.}
    \label{fig:overview}
\end{figure}

The goal of an image captioning model is to generate a caption $c_0$ for the target image $I_0$. In this paper, we focus on distinctive image captioning, which could not only describe the target image $I_0$ correctly but also include as many details as possible to distinguish $I_0$ from other similar images $\{I_1, ..., I_K\}$, where $K$ denotes the number of similar images.

% We define the objective of distinctive image captioning as follows: given a target image $I_0$, the model should generate a sentence $c_0$ that not only describe the information in human language (e.g. English) correctly but also include as many details as possible to distinguish the target image from other similar image.

% \jimmycmd{Use notations as most as possible}

Following the goal of generating distinctive captions, we first propose the R@K-CLIP metric which uses the language-vision pre-training model CLIP \cite{radford2021learning} to quantify the distinctiveness of target caption. We then propose a series of metrics called Group Embedding Gap (GEG) including GEG-Avg and GEG-Min that compare the similarity between images and captions, and use these metrics for both training and evaluation. We design a weighted GEG reward which is used in self-critical sequence training (SCST) to guide distinctive training. Note that the GEG training strategy is ``plug and play'' and can be used on any sequence prediction models.
\subsection{Similar Image Group}

We start by introducing the construction process of the similar image groups $\{I_1, ..., I_K\}$. In each split of the training, validation and test dataset, we construct the similar image group within the same split. The $K$ similar images are retrieved according to the semantic similarity between their ground truth captions and the target image $I_0$ by image-to-text retrieval.  

Different from \cite{wang2020compare} which uses VSE++ \cite{faghri2017vse++} pre-trained model, we use CLIP (Contrastive Language-Image Pre-Training) \cite{radford2021learning} trained on larger dataset to perform image-to-text retrieval. CLIP is a neural network trained on a 400 million of image-text pairs collected from a variety of publicly available sources on the Internet. CLIP contains an image encoder $E_i(I)$ and a text encoder $E_t(c)$, which takes $I$ and $c$ as the input image and text, and output their embeddings respectively. In this paper, we use the ViT-B/32 architecture, of which the image encoder $E_i(\cdot)$ is a vision transformer with 12 transformer layers. The architecture of this vision transformer has minor change compared with the original version as an layer normalization is added to the combined patch and positional embeddings before the transformer. The text encoder $E_t(\cdot)$ first encode the raw text using a lower-cased byte paired encoding (BPE) with 49,152 vocab size and then fed them to a Transformer also with 12 transformer layers. Experimental result in section \ref{sec:main_results} demonstrates that CLIP has better image retrieval performance than VSE++.The similarity of image $I$ and caption $c$ is defined as the inner product of their encoding representations:
\begin{align}
    f_{\scalebox{.5}{CLIP}}(I, c) &= \frac{E_{i}(I)^T E_{t}(c)}
    {\left\lVert {E_{i}(I)} \right\rVert \left\lVert{E_{t}(c)} . \right\rVert}
\end{align}

Note that CLIP, as well as other cross-modal retrieval models are not trained on image-image pairs. Therefore, to obtain the semantic similarity between images, we define the similarity score of target images $I_0$ and candidate similar image $I_i$ through the ground-truth captions of $I_i$ as following:
\begin{align}
    s(I_0, I_i) &= \max_{n \in \{1, 2, \dots, N\}}f_{\scalebox{.5}{CLIP}}(I_0, c_i^n) \,,
\end{align}
where $N$ is number of ground-truth captions for each image and $c_i^n \in G_t(I_i)$ is the $n$-th ground-truth caption of image $I_i$.

In practice, we first construct the set of similar captions $\{c_1', c_2', ..., c_{N'}'\}$, where $N' = N(K+1)$ and $K$ is the size of similar image group. In this way, at least $K$ different images corresponding to the captions in the similar caption set are included. Then we choose the top $K$ images to form the similar image group $S_{im}(I_0)$.

The similar image groups for all images in each split of the dataset are generated offline. The training process uses similar image groups in both forward and backward propagation. However, even though the evaluation of distinctiveness of captions involves their similar image groups, the model is unaware of the similar image groups and only takes target images as input in the test set at inference time. Therefore, we claim that similar image groups supervise the model to learn distinctive features of images in a contrastive way.

\subsection{Metrics}

\subsubsection{R@K-CLIP}

The distinctiveness of image captioning is hard to quantify because it depends on the dataset as well as the retrieval model. A general idea of evaluating distinctiveness is that the distinct caption should be able to retrieve its target image over a bunch of images via a text-image retrieval model. Previous work \cite{wang2020compare,wang2021group} use VSE++ to the retrieve images. In this paper, we choose CLIP as the retrieval model because it is pre-trained on larger dataset thus more reliable. User study in section \ref{sec:user_study} suggests that our metric for distinctivness is highly consistent with human judgement.

In this paper, we fix the retrieval model to be CLIP ViT-B/32 and evaluate the distinctiveness of the generated captions by computing the recall rate of captions that retrieves their corresponding images, which is denoted as Recall at K (R@K). In specific, we define our proposed retrieval metric R@K-CLIP as the percentage of captions which retrieves their corresponding images within the top $K$ images by CLIP ViT-B/32 model.
% \begin{align}
%     \text{R@K-CLIP}(S) = \frac{\left\lVert R_K(S) \right\rVert}{\left\lVert S \right\rVert} \,,
% \end{align}
% where
% \begin{align}
%     R_K(S) = \{I \in S \, | \, \exists I_1', ..., I_{K-1}', f_{\scalebox{.5}{CLIP}}(I, c) \} \,.
% \end{align}

\subsubsection{Group Embedding Gap}

We propose the Group Embedding Gap (GEG) metrics that can be used for both training and evaluating a distinctive captioning model. These metrics are group specified, which means they depend on the choice of similar image group for each target image. Intuitively, a distinctive caption $C$ should be close to the target image $I$ while be distant to other similar images $S_{im}(I)$. To this end, the GEG metrics measure the distinctiveness of generated caption $c$ for target image $I$ by subtracting the similarity of $c$ and $S_{im}(I)$ from the similarity of $c$ and $I$. The difference is GEG-Avg ($G_{avg}$) computes the average similarity of $c$ and every similar images in $S_{im}(I)$ while GEG-Min ($G_{min}$) focuses on the most similar one in $S_{im}(I)$:
% \wenjia{Use two sentence to introduce why you propose this metric, and what is the effect of this metric.}
\begin{align}
    G_{avg}(I, c) &= f_{\scalebox{.5}{CLIP}}(I, c) - \frac{1}{\left\lVert S_{im}(I) \right\lVert} \sum\limits_{I' \in S_{im}(I)} f_{\scalebox{.5}{CLIP}}(I', c) \,, \\
    G_{min}(I, c) &= f_{\scalebox{.5}{CLIP}}(I, c) - \max_{I' \in S_{im}(I)} f_{\scalebox{.5}{CLIP}}(I', c) \,,
\end{align}
where the CLIP model $f_{\scalebox{.5}{CLIP}}$ is used to measure the similarity between captions and images.

% \jimmycmd{better use $G_{avg}(I, c)$ and $G_{min}(I, c)$}

% \wenjia{explain the metric here. What is $\text{SIM(I)}$? What is $\text{Avg(I,c)}$?}

\subsection{Group Embedding Gap Reward}

Usually the training schedule for caption models involves both maximum likelihood estimation (MLE) training and self-critical sequence training (SCST) \cite{rennie2017self}. The purpose of SCST is to optimize a non-differentiable metric, such as CIDEr \cite{vedantam2015cider}. SCST treats the generation of captions as a reinforcement learning problem, where the expected reward of non-differentiable metric is optimized by evaluating policy gradient.

Following the sketch of SCST, we compute the policy gradient as follows:
\begin{align}
    \bigtriangledown L(\theta) = -\mathbb{E}_{c \sim P(c|I; \theta)}[r(c) - r(c_{greedy}) \bigtriangledown_{\theta} \log P(c|I; \theta)] \,,
\end{align}
where $P(c|I; \theta)$ is the probability distribution of the model generating caption $c$ for target image $I$ with model parameter $\theta$. $c_{greedy}$ is the caption generated via greedy decoding, i.e. choosing the word with highest probability at every time step. The expectation of the gradient of loss can be estimated by sampling $c$ using beam search. Therefore, for each sampled caption $c_{beam}$,
\begin{align}
    \bigtriangledown L(\theta) = -(r(c_{beam}) -   r(c_{greedy})) \bigtriangledown_{\theta} \log P(c_{beam}|I; \theta) \,.
\end{align}
$r(c)$ originally take the CIDEr score of caption $c$ as reward. Here we add the group embedding gap metric as a new component and re-weight the original CIDEr score reward and group embedding gap reward to get the final reward for sampled caption $c$:
\begin{align}
    r(c) = \alpha \text{CIDEr}(c, \hat{c}) + \beta G(I, c) \,,
\end{align}
where $G$ represents the choice of group embedding gap metric of either $G_{avg}$ or $G_{min}$. $\alpha$ and $\beta$ are hyper-parameters to control weights of the two rewards. 

\section{Experiments}

In this section, we first introduce our experiment settings. Then we validate the reliability of the R@K-CLIP metrics. Next we evaluate the improvement of our group based optimization strategy as for distinctiveness on a variety of captioning models, and compare the best model with existing state-of-the-art. Finally, user study and qualitative results are provided to show the consistency of metric based results and the human evaluation.

\subsection{Implementation Details}

\subsubsection{Dataset}

All our experiments are performed on MSCOCO dataset \cite{lin2014microsoft} with Karpathy split \cite{karpathy2015deep}, which consists of a training set with 113,287 images, validation set with 5,000 images, and test set with 5,000 images. Each image is annotated with 5 captions. In this paper, all experiments results are reported on the test set of Karpathy split.

\subsubsection{Models}

\label{sec:models}

Following \cite{li2021x}, we evaluate our optimization strategy on seven different baseline model architectures: LSTM-A3 \cite{yao2017boosting}, Attention \cite{xu2015show}, Up-Down \cite{anderson2018bottom}, GCN-LSTM \cite{yao2018exploring}, Transformer \cite{sharma2018conceptual}, Meshed-Memory \cite{cornia2020meshed} and X-LAN \cite{pan2020x}.

Two types of features, global features and top-down features are used in these models. Global features are extracted from Resnet-101 \cite{he2016deep} and top-down features are spatial features extracted from Faster-RCNN \cite{ren2015faster}. For each image, the global feature is a vector of dimension 2,048 and top-down feature is a matrix of shape $N_{RoI} \times 1,024$ where $N_{RoI}$ is the number of regions of interest. In specific, global features are used by LSTM-A3. Top-down features are used by Attention, Up-down, GCN-LSTM, Transformer, Meshed-Memory and X-LAN.

Each model is trained using three methods: 1) standard MLE objective with cross-entropy loss, denoted as ``model"; 2) SCST optimization, which is first trained with MLE objective then optimized by SCST by CIDEr reward, denoted as ``model+SCST"; 3) GEG optimization, which uses SCST to optimize the weighted group embedding gap reward based on the SCST model, and is denoted as ``model+SCST+GEG".

\subsubsection{Training Details}

\label{sec:training_details}

We apply GEG reward training on each model after SCST optimization. For GEG reward training, we collect $K = 5$ images to form the similar image group and set $\alpha = 0.1$ and $\beta = 10$. Maintaining a non-zero $\alpha$ is crucial because CLIP is not trained with language modeling objective. Ablation study in section \ref{sec:ablation_study} shows that training using only GEG reward results in grammarly incorrect captions. We use Adam optimizer with base learning rate $r = 5 \times 10^{-5}$, $\beta_1 = 0.9$, $\beta_2 = 0.999$ and $\epsilon = 10^{-8}$. Each model is trained with GEG reward for 10 epochs.

\subsubsection{Metrics}

We use two groups of metrics. The first group includes language quality metrics, i.e., CIDEr, BLEU@1, BLEU@2, BLEU@3, BLEU@4, METEOR, ROUGE-L and SPICE, to evaluate the similarity of generated captions and ground-truth captions. The second group includes distinctiveness metrics of retrieval metrics, i.e.,  R@1-CLIP, R@5-CLIP, R@10-CLIP and GEG metrics GEG-Avg, GEG-Min.

\label{sec:metrics_evaluation}
We report the image retrieval performance of CLIP ViT-B/32 on MSCOCO test set in Table~\ref{tab1}. The results suggest that CLIP ViT-B/32 has generally better performance than VSE++ because of higher R@1 and R@10 and slightly lower R@5.

\begin{table}[pbt]
\centering
\begin{tabular}{l|cccc}
\cline{1-4}
Model     & R@1  & R@5  & R@10 &  \\ \cline{1-4}
VSE++ \cite{faghri2017vse++}   & 24.1 & \textbf{52.8} & 66.2 &  \\ \cline{1-4}
CLIP \cite{radford2021learning}      & \textbf{30.4} & 50.0 & \textbf{66.9} &  \\ \cline{1-4}
\end{tabular}
\vspace{0.2cm}
\caption{Image Retrieval performance of different models on MS COCO Karpathy test split.}
\label{tab1}
\end{table}

\subsection{Main Results}

\label{sec:main_results}
We report the main results of applying different training strategies on seven baseline captioning models in Table \ref{tab:metric_results}. The seven models are ranked by CIDEr score of ``model+SCST" from low to high. In specific, we consider CIDEr as the main language quality metric to measure the similarity of generated caption and ground-truth captions. The remaining language quality metrics are positively correlated to CIDEr. We also consider R@K-CLIP metrics as the main distinctiveness metrics, and GEG metrics are positively correlated to R@K-CLIP.

By observing the results in Table \ref{tab:metric_results}, we have the following observations. Firstly, by comparing the models trained by MLE or SCST from top to bottom, the results suggest that models with higher language quality metrics tend to have higher distinctiveness metrics. For instance, compared to Up-Down~\cite{anderson2018bottom} getting CIDEr value 113.1 and R@1-CLIP at 16.08, GCN-LSTM \cite{yao2018exploring} achieves higher CIDEr at 116.3 and R@1-CLIP at 17.46. This correlation implies the consistency of the two types of metrics. When no extra supervision or objective is provided, models which can better mimic human annotations describes more distinctive details of the image. Then, by comparing ``model+SCST+GEG" with the baseline ``model+SCST", the results suggest that models trained by GEG reward generate much more distinctive captions. For instance, compared to X-LAN+SCST getting R@K-CLIP rates at 21.3, 44.74 and 55.74 repectively, X-LAN+SCST+GEG achieves higher R@K-CLIP rates at 28.12, 50.3, 67.18 respectively. This improvement demonstrates the effectiveness of our proposed methods. Although the language metrics get lower, for instance, in the X-LAN group, X-LAN+SCST+GEG has lower CIDEr (121.7) than X-LAN+SCST (130.0), we argue that this is because human annotated ground-truth captions only consider the consistency with target images and are not distinctive in many cases. We provide case study in section \ref{sec:qualitative_results} to demonstrate that the decrease in language metrics does not negatively affect caption quality and on the contrary generates better captions than both baseline captions and ground-truth captions 
% \wenjia{How about GEC-Avg and GEC-Min? What conclusion for these two?}

\begin{table}[pbt]
\centering
\resizebox{340pt}{105pt}{
\begin{tabular}{l|cccc|ccccccccc}
\toprule
Model                  & CIDEr  & R@1 & R@5 & R@10 & B@1 & B@2 & B@3 & B@4 & M & R & S & GEG-Avg & GEG-Min \\ \hline
LSTM-A3 \cite{yao2017boosting}               & 107.7  & 11.52    & 30.88    & 42.34     & 75.3   & 59.0   & 45.4   & 35.0   & 26.7    & 55.6    & 19.7  & 0.0296  & -0.0064 \\
LSTM-A3+SCST           & 117.0  & 10.98    & 29.38    & 41.34     & 77.9   & 61.5   & 46.7   & 35.0   & 27.1    & 56.3    & 20.5  & 0.0288  & -0.0073 \\
LSTM-A3+SCST+GEG       & 114.3  & 12.98    & 32.86    & 45.7      & 76.9   & 60.4   & 45.6   & 33.9   & 26.9    & 55.8    & 20.6  & 0.0330  & -0.0046 \\ \hline
Attention \cite{xu2015show}             & 113.0  & 14.74    & 36.4     & 48.6      & 76.4   & 60.6   & 46.9   & 36.1   & 27.6    & 56.6    & 20.4  & 0.0344  & -0.0015 \\
Attention+SCST         & 123.1  & 14.8     & 35.5     & 47.04     & 79.4   & 63.5   & 48.9   & 37.1   & 27.9    & 57.6    & 21.3  & 0.0338  & -0.0026 \\
Attention+SCST+GEG     & 116.6  & 17.68    & 40.94    & 52.84     & 77.1   & 60.8   & 45.9   & 34.3   & 27.4    & 56.4    & 21.1  & 0.0406  & 0.0026  \\ \hline
Up-Down \cite{anderson2018bottom}               & 113.1  & 16.08    & 37.4     & 49.18     & 76.3   & 60.3   & 46.6   & 36.0   & 27.6    & 56.6    & 20.7  & 0.0351  & -0.0008 \\
Up-Down+SCST           & 124.7  & 15.46    & 36.2     & 49.28     & 80.1   & 64.3   & 49.7   & 37.7   & 28.0    & 58.0    & 21.5  & 0.0349  & -0.0014 \\
Up-Down+SCST+GEG       & 114.2  & 19.78    & 42.7     & 55.68     & 76.24  & 60.15  & 45.46  & 33.85  & 27.41   & 56.24   & 21.17 & 0.0424  & 0.0048  \\ \hline
GCN-LSTM \cite{yao2018exploring}              & 116.3  & 17.46    & 40.02    & 52.68     & 76.8   & 61.1   & 47.6   & 36.9   & 28.2    & 57.2    & 21.2  & 0.0373  & 0.0011  \\
GCN-LSTM+SCST          & 127.2  & 17.22    & 39.46    & 52.16     & 80.2   & 64.7   & 50.3   & 38.5   & 28.5    & 58.4    & 22.1  & 0.0375  & 0.0015  \\
GCN-LSTM+SCST+GEG      & 112.6  & 20.66    & 45.26    & 58.16     & 76.30  & 60.04  & 45.50  & 33.74  & 26.92   & 55.76   & 20.83 & 0.0441  & 0.0063  \\ \hline
Transformer \cite{sharma2018conceptual}           & 116.6  & 18.76    & 42.26    & 54.16     & 76.4   & 60.3   & 46.5   & 35.8   & 28.2    & 56.7    & 21.3  & 0.0391  & 0.0028  \\
Transformer+SCST       & 130.0  & 19.6     & 42.72    & 55.62     & 80.5   & 65.4   & 51.1   & 39.2   & 29.1    & 58.7    & 23.0  & 0.0403  & 0.0033  \\
Transformer+SCST+GEG   & 123.2  & 25.06    & 50.8     & 62.96     & 77.66  & 61.83  & 47.57  & 35.79  & 28.58   & 57.36   & 22.84 & 0.0485  & 0.0102  \\ \hline
Meshed-Memory \cite{cornia2020meshed}         & 116.0  & 18.66    & 41.12    & 52.92     & 76.3   & 60.2   & 46.4   & 35.6   & 28.1    & 56.5    & 21.2  & 0.0383  & 0.0024  \\
Meshed-Memory+SCST     & \textbf{131.1}  & 19.64    & 42.94    & 55.86     & \textbf{80.7}   & \textbf{65.5}   & \textbf{51.4}   & 39.6   & 29.2    & 58.9    & 22.9  & 0.0402  & 0.0034  \\
Meshed-Memory+SCST+GEG & 124,4  & 25.44    & 50.88    & 63.7      & 78.15  & 62.33  & 48.17  & 36.48  & 28.74   & 57.48   & 22.91 & 0.0484  & 0.0102  \\ \hline
X-LAN \cite{pan2020x}                 & 120.7  & 20.26    & 43.88    & 55.4      & 77.5   & 61.9   & 48.3   & 37.5   & 28.6    & 57.6    & 21.9  & 0.0402  & 0.0036  \\
X-LAN+SCST             & 130.0  & 21.3     & 44.74    & 57.74     & 80.4   & 65.2   & 51.0   & 39.2   & \textbf{29.4}    & \textbf{59.0}    & \textbf{23.2}  & 0.0416  & 0.0045  \\
X-LAN+SCST+GEG         & 121.7  & \textbf{28.12}    & \textbf{50.3}     & \textbf{67.18}     & 76.23  & 60.55  & 46.37  & 34.82  & 28.66   & 56.96   & 22.84 & \textbf{0.0517}  & \textbf{0.0127}  \\ \hline
Stack-Cap \cite{gu2018stack}              & 120.4  & 13.96    & 34.36    & 45.78     & -      & -      & 47.9   & 36.1   & 27.4    & 56.9    & 20.9  & 0.0339  & -0.0019 \\
Disc-Cap \cite{luo2018discriminability}               & 120.1  & 11.24    & 29.74    & 41.86     & -      & -      & 48.5   & 36.1   & 27.7    & 57.8    & 21.4  & 0.0313  & -0.0042 \\
CIDErBtw-Cap \cite{wang2020compare}           & 127.9 & 19.16    & 41.98    & 54.54     & -      & -      & 51.0   & 38.5   & 29.1    & 58.2    & 23.0  & 0.0390  & 0.0025  \\
Gdis-Cap \cite{wang2021group}               & 127.5  & 17.62    & 38.94    & 51.68     & -      & -      & 50.0   & 38.1   & -       & -       & -     & -       & -       \\ \bottomrule
\end{tabular}
}
\vspace{0.2cm}
\caption{The main results of applying different training strategy, i.e., the MLE optimization~(denoted as ``model"), the SCST optimization~(denoted as ``model + SCST"), and the GEG optimization~(denoted as ``+ GEG"), on seven baseline models. Here R@K represents R@K-CLIP and B@K represents BLEU@K. M, R and S represent METEOR, ROUGE-L and SPICE respectively.}
\label{tab:metric_results}
\end{table}

% \wenjia{add reference. use abbrievation}

\subsection{Comparison with State-of-the-art}

We compare our model with best performance X-LAN+SCST+GEG with three state-of-the-art captioning models which also focus on distinctiveness object at the bottom of Table \ref{tab:metric_results}. We compare distinctiveness of each model based on R@K-CLIP and GEG metrics. Compared with Stack-Cap \cite{gu2018stack} and Disc-Cap \cite{luo2018discriminability}, our model shows better performance in both language quality metrics and distinctiveness metrics. For instance, comparing to Stack-Cap \cite{gu2018stack} which has CIDEr at 120.4 and R@1-CLIP at 13.96, our X-LAN+SCST+GEG has higher CIDEr at 121.7 and higher R@1-CLIP at 28.12. Compared with CIDErBtw-Cap \cite{wang2020compare} and Gdis-Cap \cite{wang2021group}, our model has lower CIDEr score but still much better performance in distinctiveness . Recall that decrease in CIDEr score does not imply the reduction of language quality. Therefore, we claim that our model achieves new state-of-the-art towards distinctive captioning.

\subsection{Ablation Study}

\label{sec:ablation_study}

In order to further illustrate the effectiveness of our training strategy. We provide an ablation study including five experiments as shown in Table \ref{tab:ablation_study}. We compare our proposed standard SCST+GEG strategy with four other strategies on X-LAN model. X-LAN+SCST is the same baseline model trained using SCST as in Table \ref{tab:metric_results}. X-LAN+SCST++ is the baseline model trained for 10 more epochs so X-LAN+SCST++ and X-LAN+SCST+GEG are trained for the same number of epochs. Instead of using weighted GEG reward, X-LAN+SCST+ER only takes the CLIP embedding similarity of generated caption and target image as reward. In this case, the embedding reward does not include the negative average similarity of generated caption and similar images. X-LAN+SCST+GEG-sole sets $\alpha = 0$ which removes CIDEr reward and only takes GEG reward.

By analysing Table \ref{tab:ablation_study}, we have the following observations. Firstly, the results of comparing X-LAN+SCST++ and X-LAN+SCST shows that further training has little improvement on both CIDEr metric and distinctiveness metric. Introducing the possitive similarity of generated caption and the target image as reward (i.e., X-LAN+SCST+ER) already helps the model in improving the distinctivess. For instance, X-LAN+SCST+ER improves the R@1-CLIP score from 21.8~(of X-LAN+SCST) to 25.86. Comparison between X-LAN+SCST+GEG and X-LAN+SCST+ER shows the effectiveness of group based learning. By negatively rewarding the similarity of generated caption and similar images, the X-LAN+SCST+GEG model generates captions that are better in both language quality metric (e.g., improves CIDEr from 120.7 to 121.7) and distinctive metric~(e.g., promoting R@1-CLIP from 25.86 to 28.13).  X-LAN+SCST+GEG-sole model has unexpectedly high distinctiveness metric without the CIDEr score constraint. However, as discussed in section \ref{sec:training_details}, CLIP is not trained with language quality objective. Training with only GEG reward causes frequent grammar mistakes of word disorder and repentance. An example is ``a woman eating a white dessert cake with lit candles with lit candles in a restaurant with lit candles with", which is completely unreadable.

\begin{table}[pbt]
\centering
\begin{tabular}{l|cccc}
\toprule
Model               & R@1-CLIP & R@5-CLIP & R@10-CLIP & CIDEr \\ \hline

X-LAN+SCST          & 21.3     & 44.74    & 57.74     & 130.0 \\
X-LAN+SCST++        & 21.8     & 44.24    & 57.98     & \textbf{130.5} \\
X-LAN+SCST+ER       & 25.86    & 52.38    & 64.88     & 120.7 \\ 
X-LAN+SCST+GEG  (Full model)    & \textbf{28.13}    & \textbf{50.3}     & \textbf{67.18}     & 121.7 \\ \hline
X-LAN+SCST+GEG-sole & 32.82    & 61.3     & 72.8      & 17.4  \\ \hline
\end{tabular}
\vspace{0.2cm}
\caption{Comparison between different training strategies}
\label{tab:ablation_study}
\end{table}

\begin{table}[pbt]
\centering
\begin{tabular}{l|ccc}
\toprule
Model          & R@1-User & R@5-User & R@10-User \\ \hline
X-LAN+SCST     & 21.9     & 46.3     & 65.4      \\
X-LAN+SCST+GEG & \textbf{31.7}     & \textbf{55.7}     & \textbf{71.5}      \\ \bottomrule
\end{tabular}
\vspace{0.2cm}
\caption{User-evaluated image retrieval}
\label{tab:user_study_1}
\end{table}

\begin{table}[pbt]
\centering
\begin{tabular}{l|ccc}
\toprule
Criterion       & Win(\%)  & Tie(\%)  & Loss(\%) \\ \hline
Accuracy        & 16.7 & 80.3 & 3.0  \\
Distinctiveness & 44.5 & 47.2 & 8.3  \\ \bottomrule
\end{tabular}
\vspace{0.2cm}
\caption{Comparison on accuracy and distinctiveness. The rates of win, tie, loss are related to X-LAN+SCST+GEG model.}
\label{tab:user_study_2}
\end{table}

\begin{table}[pbt]
\centering
\resizebox{340pt}{250pt}{
\begin{tabular}{llllll}
\toprule
\multicolumn{6}{c}{\includegraphics[width=1.25\linewidth]{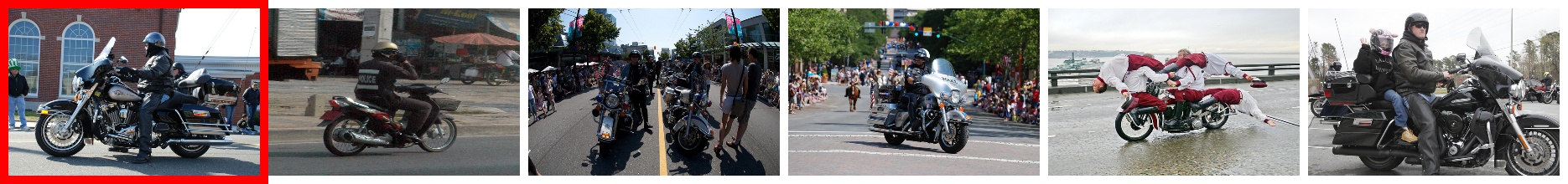}}                                                                                                                                              \\ \hline
\multicolumn{1}{l|}{X-LAN+SCST}                            & \multicolumn{5}{l}{A man sitting on a motorcycle in the street.}                                     \\
\multicolumn{1}{l|}{X-LAN+SCST+GEG}                        & \multicolumn{5}{l}{A man sitting on a motorcycle in front of \textcolor{red}{a brick building}.}                      \\ \hline
\multicolumn{1}{l|}{\multirow{5}{*}{Groud-Truth Captions}} & \multicolumn{5}{l}{A couple of men that are on some motorcycles.}                                    \\
\multicolumn{1}{l|}{}                                      & \multicolumn{5}{l}{A cop riding on the back of a black motorcycle.}                                  \\
\multicolumn{1}{l|}{}                                      & \multicolumn{5}{l}{A man on a motorcycle in a parade.}                                               \\
\multicolumn{1}{l|}{}                                      & \multicolumn{5}{l}{A motorcycle rider dressed in leather drives a motorcycle down a crowded street.} \\
\multicolumn{1}{l|}{}                                      & \multicolumn{5}{l}{A man on a motorcycle stopped on a city street.}                                  \\ \hline \hline
\multicolumn{6}{c}{\includegraphics[width=1.25\linewidth]{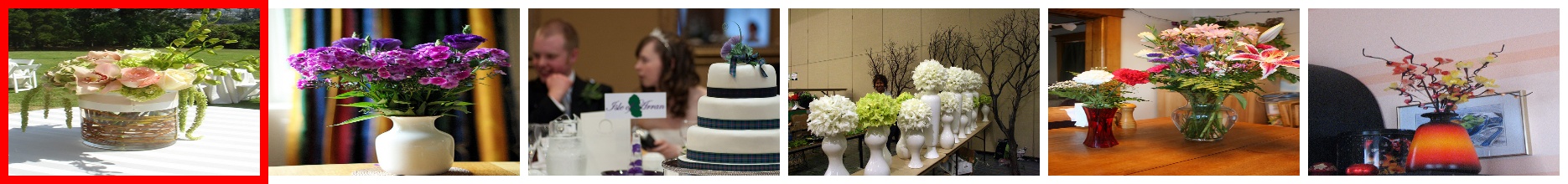}}                                                                                                                                              \\ \hline
\multicolumn{1}{l|}{X-LAN+SCST}                            & \multicolumn{5}{l}{A vase filled with flowers on a table.}                                           \\
\multicolumn{1}{l|}{X-LAN+SCST+GEG}                        & \multicolumn{5}{l}{A \textcolor{red}{glass} vase filled with \textcolor{red}{pink roses} on a table.}                                  \\ \hline
\multicolumn{1}{l|}{\multirow{5}{*}{Groud-Truth Captions}} & \multicolumn{5}{l}{Creative centerpiece floral arrangement at an outdoor event.}                     \\
\multicolumn{1}{l|}{}                                      & \multicolumn{5}{l}{A wedding centerpiece made of flowers and various other plants.}                  \\
\multicolumn{1}{l|}{}                                      & \multicolumn{5}{l}{A vase of flowers sitting on an outdoor table.}                                   \\
\multicolumn{1}{l|}{}                                      & \multicolumn{5}{l}{A vase filled with flowers on top of a table.}                                    \\
\multicolumn{1}{l|}{}                                      & \multicolumn{5}{l}{A floral arrangement inside a clear cylinder shaped vase.}                        \\ \hline \hline
\multicolumn{6}{c}{\includegraphics[width=1.25\linewidth]{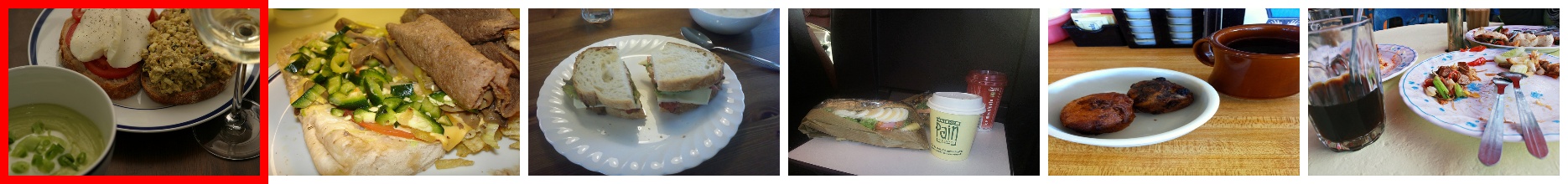}}                                                                                                                                              \\ \hline
\multicolumn{1}{l|}{X-LAN+SCST}                            & \multicolumn{5}{l}{A plate with a sandwich and a bowl of soup.}                                      \\
\multicolumn{1}{l|}{X-LAN+SCST+GEG}                        & \multicolumn{5}{l}{A sandwich with \textcolor{red}{meat and tomatoes} on a \textcolor{red}{white plate} next to a bowl of soup.}       \\ \hline
\multicolumn{1}{l|}{\multirow{5}{*}{Groud-Truth Captions}} & \multicolumn{5}{l}{A plate with an open sandwich next to a little bowl of soup and a wine glass.}    \\
\multicolumn{1}{l|}{}                                      & \multicolumn{5}{l}{A plate topped with an open face sandwich net to a glass of wine.}                \\
\multicolumn{1}{l|}{}                                      & \multicolumn{5}{l}{There is a plate of food and a bowl of soup.}                                     \\
\multicolumn{1}{l|}{}                                      & \multicolumn{5}{l}{A plate with a sandwich, a bowl of soup and a wine glass.}                        \\
\multicolumn{1}{l|}{}                                      & \multicolumn{5}{l}{A plate of food that has an open faced sandwich on it.}                           \\ \hline \hline
\multicolumn{6}{c}{\includegraphics[width=1.25\linewidth]{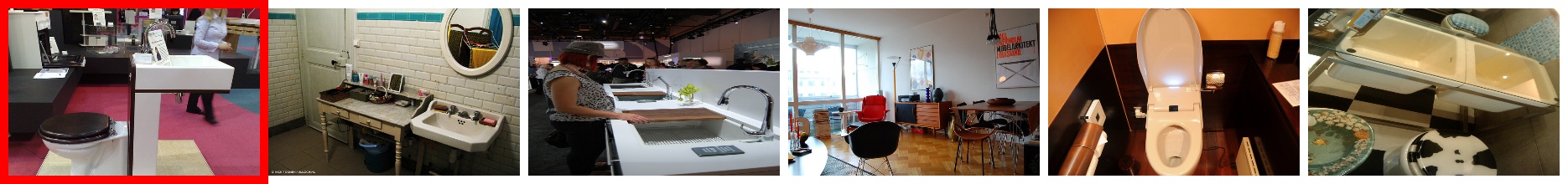}}                                                                                                                                              \\ \hline
\multicolumn{1}{l|}{X-LAN+SCST}                            & \multicolumn{5}{l}{A bathroom with a toilet and a sink.}                                             \\
\multicolumn{1}{l|}{X-LAN+SCST+GEG}                        & \multicolumn{5}{l}{\textcolor{red}{A woman} standing next to a toilet in \textcolor{red}{a store.}}                                    \\ \hline
\multicolumn{1}{l|}{\multirow{5}{*}{Groud-Truth Captions}} & \multicolumn{5}{l}{A display of a toilet and sink in a store.}                                       \\
\multicolumn{1}{l|}{}                                      & \multicolumn{5}{l}{Toilet and sink on display in a furniture store.}                                 \\
\multicolumn{1}{l|}{}                                      & \multicolumn{5}{l}{A toilet with a sink on the back is displayed.}                                   \\
\multicolumn{1}{l|}{}                                      & \multicolumn{5}{l}{A store filled with bathroom equipment and accessories.}                          \\
\multicolumn{1}{l|}{}                                      & \multicolumn{5}{l}{A white toilet with a brown lid and a sink behind it.}                            \\ \hline \hline
\multicolumn{6}{c}{\includegraphics[width=1.25\linewidth]{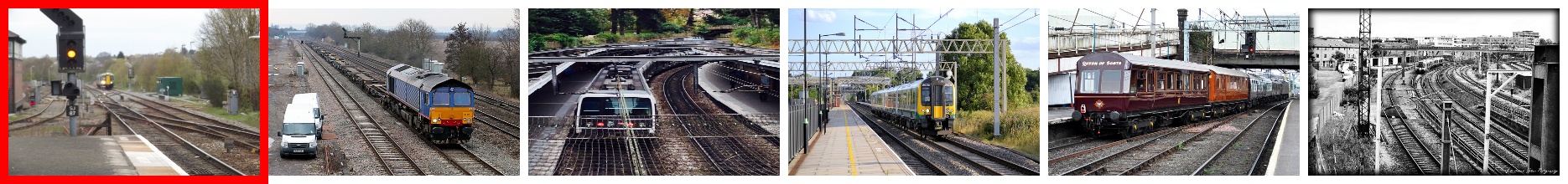}}                                                                                                                                              \\ \hline
\multicolumn{1}{l|}{X-LAN+SCST}                            & \multicolumn{5}{l}{A yellow train is traveling down the tracks.}                                     \\
\multicolumn{1}{l|}{X-LAN+SCST+GEG}                        & \multicolumn{5}{l}{A yellow train traveling down the tracks next to \textcolor{red}{a traffic light.}}                \\ \hline
\multicolumn{1}{l|}{\multirow{5}{*}{Groud-Truth Captions}} & \multicolumn{5}{l}{A train is near a track switching station.}                                       \\
\multicolumn{1}{l|}{}                                      & \multicolumn{5}{l}{A train on a track near many trees.}                                              \\
\multicolumn{1}{l|}{}                                      & \multicolumn{5}{l}{A train crossing with a train coming down the railroad tracks.}                   \\
\multicolumn{1}{l|}{}                                      & \multicolumn{5}{l}{The train is slowing its speed heading into the station.}                         \\
\multicolumn{1}{l|}{}                                      & \multicolumn{5}{l}{A yellow light is next to some train tracks.}                                     \\ \bottomrule
\end{tabular}
}
\vspace{0.2cm}
\caption{Example captions from the baseline model and our model with similar images. The leftmost images with red borders are the target images. Red words show the distinctive details in the target images which are not mentioned in either baseline model captions or ground-truth captions}
\label{tab:qualitative_results}
\end{table}

\subsection{User Study}

\label{sec:user_study}

In order to fairly evaluate the improvement of our model and verify the consistency between our proposed metrics and human perspective, we conducted two user studies.

In the first experiment, each user is given a random group of images and a caption. The caption is generated from either the baseline X-LAN+SCST model or X-LAN+SCST+GEG model. The group of images contains an target image corresponding to the caption and 20 similar images. The user is asked to list in order the 10 most relevant images. In this way, we obtain the R@K-User metric.

In the second experiment, each user is asked to give preference of two captions generated by the above two models following two criteria: accuracy, which caption describes the target image more accurately; distinctiveness, which caption is more informative about the target image. The user can choose win, loss or tie of the X-LAN+SCST+GEG model.

We received 1,200 responses in total. The results are shown in Table \ref{tab:user_study_1} and Table \ref{tab:user_study_2}. Compared with baseline model X-LAN+SCST, our model increases the image retrieval rate R@1-User, R@5-User and R@10-User by 9.8, 9.4 and 6.1, and surpasses the baseline model in both accuracy and distinctiveness. Both observations show that human evaluation is consistent with metric results in section \ref{sec:main_results}.

\subsection{Qualitative Results}

\label{sec:qualitative_results}

The results shown in Table \ref{tab:qualitative_results} illustrates the improvement of distinctiveness of our model compared with baseline model. Due to the space limit, we only show five groups of results and analyze the first two groups.

In each group of the images, the leftmost image is the target image and followed by five similar images. We compare the generated captions from the baseline model X-LAN+SCST and our model X-LAN+SCST+GEG. In both cases, the baseline model generates accurate but generic captions that describe the salient object in the target image. Contrastively, our model describes more details that distinguish the target image from potential similar images. In the first case, our model not only captures the main object, i.e. ``man" and ``motorcycle", but also notice the ``brick building" in the background. In the second case, our model extends the generic description of ``vase" and ``flowers" to more specific description ``glass vase" and ``pink roses". Recall that during inference, our model is unaware of the similar images in test set. In this case, after distinctive training, our model is able to capture the details that distinguish the target image from potential similar images, which demonstrate the effectiveness of our training strategy.

The two cases also explain the decrease in language quality metrics, e.g. CIDEr. In the first case, none of the ground-truth captions mentions ``brick building". In the second case, none of the ground-truth captions mentions ``glass vase" and ``pink roses". However, these ignored details are crucial to distinguish the target images. Training with only language quality objective results in generic captions possibly because the model is stuck in the minimum which is close to every ground-truth captions. After distinctive training, our model can discover more details than human annotated captions, which is considered as the main reason of the decrease in CIDEr metric.

% \wenjia{Use red box to mark out the target image.}

\section{Conclusions}

In this paper, we focus on the distinctiveness of image captioning. We propose the R@K-CLIP metrics and GEG metrics to evaluate distinctiveness of captioning models. In order to improve distinctiveness, we design a simple and effective strategy which optimize on similar image group to guide distinctive training. In this process, the large-scale vision-language pre-trained model CLIP is used to measure the similarity of image-caption pairs. We conduct extensive experiments on various existing models and demonstrate the wide applicability and effectiveness of the group based optimization strategy in term of improving distinctiveness. Finally, we provide qualitative results and user study to show the consistency of the metric-side improvement with human judgement.

% \clearpage\mbox{}Page \thepage\ of the manuscript.
% \clearpage\mbox{}Page \thepage\ of the manuscript.

% This is the last page of the manuscript.
% Now we have reached the maximum size of the ECCV 2022 submission (excluding references).
% References should start immediately after the main text, but can continue on p.15 if needed.

\clearpage
\bibliographystyle{splncs04}
\bibliography{egbib}

\begin{thebibliography}{10}
\providecommand{\url}[1]{\texttt{#1}}
\providecommand{\urlprefix}{URL }
\providecommand{\doi}[1]{https://doi.org/#1}

\bibitem{anderson2016spice}
Anderson, P., Fernando, B., Johnson, M., Gould, S.: Spice: Semantic
  propositional image caption evaluation. In: European conference on computer
  vision. pp. 382--398. Springer (2016)

\bibitem{anderson2018bottom}
Anderson, P., He, X., Buehler, C., Teney, D., Johnson, M., Gould, S., Zhang,
  L.: Bottom-up and top-down attention for image captioning and visual question
  answering. In: Proceedings of the IEEE conference on computer vision and
  pattern recognition. pp. 6077--6086 (2018)

\bibitem{banerjee2005meteor}
Banerjee, S., Lavie, A.: Meteor: An automatic metric for mt evaluation with
  improved correlation with human judgments. In: Proceedings of the acl
  workshop on intrinsic and extrinsic evaluation measures for machine
  translation and/or summarization. pp. 65--72 (2005)

\bibitem{bengio2015scheduled}
Bengio, S., Vinyals, O., Jaitly, N., Shazeer, N.: Scheduled sampling for
  sequence prediction with recurrent neural networks. Advances in neural
  information processing systems  \textbf{28} (2015)

\bibitem{cho2022fine}
Cho, J., Yoon, S., Kale, A., Dernoncourt, F., Bui, T., Bansal, M.: Fine-grained
  image captioning with clip reward. arXiv preprint arXiv:2205.13115  (2022)

\bibitem{cho2015describing}
Cho, K., Courville, A., Bengio, Y.: Describing multimedia content using
  attention-based encoder-decoder networks. IEEE Transactions on Multimedia
  \textbf{17}(11),  1875--1886 (2015)

\bibitem{cho2014learning}
Cho, K., Van~Merri{\"e}nboer, B., Gulcehre, C., Bahdanau, D., Bougares, F.,
  Schwenk, H., Bengio, Y.: Learning phrase representations using rnn
  encoder-decoder for statistical machine translation. arXiv preprint
  arXiv:1406.1078  (2014)

\bibitem{cornia2020meshed}
Cornia, M., Stefanini, M., Baraldi, L., Cucchiara, R.: Meshed-memory
  transformer for image captioning. In: Proceedings of the IEEE/CVF conference
  on computer vision and pattern recognition. pp. 10578--10587 (2020)

\bibitem{dai2017towards}
Dai, B., Fidler, S., Urtasun, R., Lin, D.: Towards diverse and natural image
  descriptions via a conditional gan. In: Proceedings of the IEEE international
  conference on computer vision. pp. 2970--2979 (2017)

\bibitem{dai2017contrastive}
Dai, B., Lin, D.: Contrastive learning for image captioning. Advances in Neural
  Information Processing Systems  \textbf{30} (2017)

\bibitem{faghri2017vse++}
Faghri, F., Fleet, D.J., Kiros, J.R., Fidler, S.: Vse++: Improving
  visual-semantic embeddings with hard negatives. arXiv preprint
  arXiv:1707.05612  (2017)

\bibitem{frankel1996webseer}
Frankel, C., Swain, M.J., Athitsos, V.: Webseer: An image search engine for the
  world wide web. Tech. rep., Technical Report 96-14, University of Chicago,
  Computer Science Department (1996)

\bibitem{gu2018stack}
Gu, J., Cai, J., Wang, G., Chen, T.: Stack-captioning: Coarse-to-fine learning
  for image captioning. In: Proceedings of the AAAI Conference on Artificial
  Intelligence. vol.~32 (2018)

\bibitem{he2016deep}
He, K., Zhang, X., Ren, S., Sun, J.: Deep residual learning for image
  recognition. In: Proceedings of the IEEE conference on computer vision and
  pattern recognition. pp. 770--778 (2016)

\bibitem{hochreiter1997long}
Hochreiter, S., Schmidhuber, J.: Long short-term memory. Neural computation
  \textbf{9}(8),  1735--1780 (1997)

\bibitem{karpathy2015deep}
Karpathy, A., Fei-Fei, L.: Deep visual-semantic alignments for generating image
  descriptions. In: Proceedings of the IEEE conference on computer vision and
  pattern recognition. pp. 3128--3137 (2015)

\bibitem{li2021x}
Li, Y., Pan, Y., Chen, J., Yao, T., Mei, T.: X-modaler: A versatile and
  high-performance codebase for cross-modal analytics. In: Proceedings of the
  29th ACM International Conference on Multimedia. pp. 3799--3802 (2021)

\bibitem{lin2004rouge}
Lin, C.Y.: Rouge: A package for automatic evaluation of summaries. In: Text
  summarization branches out. pp. 74--81 (2004)

\bibitem{lin2014microsoft}
Lin, T.Y., Maire, M., Belongie, S., Hays, J., Perona, P., Ramanan, D.,
  Doll{\'a}r, P., Zitnick, C.L.: Microsoft coco: Common objects in context. In:
  European conference on computer vision. pp. 740--755. Springer (2014)

\bibitem{liu2019generating}
Liu, L., Tang, J., Wan, X., Guo, Z.: Generating diverse and descriptive image
  captions using visual paraphrases. In: Proceedings of the IEEE/CVF
  International Conference on Computer Vision. pp. 4240--4249 (2019)

\bibitem{luo2018discriminability}
Luo, R., Price, B., Cohen, S., Shakhnarovich, G.: Discriminability objective
  for training descriptive captions. In: Proceedings of the IEEE Conference on
  Computer Vision and Pattern Recognition. pp. 6964--6974 (2018)

\bibitem{ma2015multimodal}
Ma, L., Lu, Z., Shang, L., Li, H.: Multimodal convolutional neural networks for
  matching image and sentence. In: Proceedings of the IEEE international
  conference on computer vision. pp. 2623--2631 (2015)

\bibitem{makav2019new}
Makav, B., K{\i}l{\i}{\c{c}}, V.: A new image captioning approach for visually
  impaired people. In: 2019 11th International Conference on Electrical and
  Electronics Engineering (ELECO). pp. 945--949. IEEE (2019)

\bibitem{mao2014deep}
Mao, J., Xu, W., Yang, Y., Wang, J., Huang, Z., Yuille, A.: Deep captioning
  with multimodal recurrent neural networks (m-rnn). arXiv preprint
  arXiv:1412.6632  (2014)

\bibitem{mirza2014conditional}
Mirza, M., Osindero, S.: Conditional generative adversarial nets. arXiv
  preprint arXiv:1411.1784  (2014)

\bibitem{pan2020x}
Pan, Y., Yao, T., Li, Y., Mei, T.: X-linear attention networks for image
  captioning. In: Proceedings of the IEEE/CVF conference on computer vision and
  pattern recognition. pp. 10971--10980 (2020)

\bibitem{papineni2002bleu}
Papineni, K., Roukos, S., Ward, T., Zhu, W.J.: Bleu: a method for automatic
  evaluation of machine translation. In: Proceedings of the 40th annual meeting
  of the Association for Computational Linguistics. pp. 311--318 (2002)

\bibitem{radford2021learning}
Radford, A., Kim, J.W., Hallacy, C., Ramesh, A., Goh, G., Agarwal, S., Sastry,
  G., Askell, A., Mishkin, P., Clark, J., et~al.: Learning transferable visual
  models from natural language supervision. In: International Conference on
  Machine Learning. pp. 8748--8763. PMLR (2021)

\bibitem{ranzato2015sequence}
Ranzato, M., Chopra, S., Auli, M., Zaremba, W.: Sequence level training with
  recurrent neural networks. arXiv preprint arXiv:1511.06732  (2015)

\bibitem{ren2015faster}
Ren, S., He, K., Girshick, R., Sun, J.: Faster r-cnn: Towards real-time object
  detection with region proposal networks. Advances in neural information
  processing systems  \textbf{28} (2015)

\bibitem{rennie2017self}
Rennie, S.J., Marcheret, E., Mroueh, Y., Ross, J., Goel, V.: Self-critical
  sequence training for image captioning. In: Proceedings of the IEEE
  conference on computer vision and pattern recognition. pp. 7008--7024 (2017)

\bibitem{sharma2018conceptual}
Sharma, P., Ding, N., Goodman, S., Soricut, R.: Conceptual captions: A cleaned,
  hypernymed, image alt-text dataset for automatic image captioning. In:
  Proceedings of the 56th Annual Meeting of the Association for Computational
  Linguistics (Volume 1: Long Papers). pp. 2556--2565 (2018)

\bibitem{shetty2017speaking}
Shetty, R., Rohrbach, M., Anne~Hendricks, L., Fritz, M., Schiele, B.: Speaking
  the same language: Matching machine to human captions by adversarial
  training. In: Proceedings of the IEEE International Conference on Computer
  Vision. pp. 4135--4144 (2017)

\bibitem{simonyan2014very}
Simonyan, K., Zisserman, A.: Very deep convolutional networks for large-scale
  image recognition. arXiv preprint arXiv:1409.1556  (2014)

\bibitem{vedantam2015cider}
Vedantam, R., Lawrence~Zitnick, C., Parikh, D.: Cider: Consensus-based image
  description evaluation. In: Proceedings of the IEEE conference on computer
  vision and pattern recognition. pp. 4566--4575 (2015)

\bibitem{vinyals2015show}
Vinyals, O., Toshev, A., Bengio, S., Erhan, D.: Show and tell: A neural image
  caption generator. In: Proceedings of the IEEE conference on computer vision
  and pattern recognition. pp. 3156--3164 (2015)

\bibitem{wang2020compare}
Wang, J., Xu, W., Wang, Q., Chan, A.B.: Compare and reweight: Distinctive image
  captioning using similar images sets. In: European Conference on Computer
  Vision. pp. 370--386. Springer (2020)

\bibitem{wang2021group}
Wang, J., Xu, W., Wang, Q., Chan, A.B.: Group-based distinctive image
  captioning with memory attention. In: Proceedings of the 29th ACM
  International Conference on Multimedia. pp. 5020--5028 (2021)

\bibitem{wang2019describing}
Wang, Q., Chan, A.B.: Describing like humans: on diversity in image captioning.
  In: Proceedings of the IEEE/CVF Conference on Computer Vision and Pattern
  Recognition. pp. 4195--4203 (2019)

\bibitem{wang2020towards}
Wang, Z., Feng, B., Narasimhan, K., Russakovsky, O.: Towards unique and
  informative captioning of images. In: European Conference on Computer Vision.
  pp. 629--644. Springer (2020)

\bibitem{xiong2019reinforced}
Xiong, Y., Du, B., Yan, P.: Reinforced transformer for medical image
  captioning. In: International Workshop on Machine Learning in Medical
  Imaging. pp. 673--680. Springer (2019)

\bibitem{xu2015show}
Xu, K., Ba, J., Kiros, R., Cho, K., Courville, A., Salakhudinov, R., Zemel, R.,
  Bengio, Y.: Show, attend and tell: Neural image caption generation with
  visual attention. In: International conference on machine learning. pp.
  2048--2057. PMLR (2015)

\bibitem{yao2018exploring}
Yao, T., Pan, Y., Li, Y., Mei, T.: Exploring visual relationship for image
  captioning. In: Proceedings of the European conference on computer vision
  (ECCV). pp. 684--699 (2018)

\bibitem{yao2017boosting}
Yao, T., Pan, Y., Li, Y., Qiu, Z., Mei, T.: Boosting image captioning with
  attributes. In: Proceedings of the IEEE international conference on computer
  vision. pp. 4894--4902 (2017)

\end{thebibliography}
\end{document}